\documentclass[runningheads]{llncs}
\usepackage[T1]{fontenc}
\usepackage{graphicx}
\usepackage{booktabs}
\usepackage{tabularx}
\usepackage{enumitem}
\usepackage{amsmath}
\usepackage{amssymb}
\usepackage{makecell} 
\usepackage{threeparttable}
\usepackage{comment}
\usepackage{xcolor}
\usepackage{tikz}
\usepackage{xcolor}
\usepackage{orcidlink}
\usepackage{wrapfig}
\usepackage{marvosym}

\usetikzlibrary{positioning, fit, arrows.meta, backgrounds}

\definecolor{ShapeTeal}{HTML}{3070B3}
\definecolor{ShapeTealFill}{HTML}{E3EEFA}
\definecolor{ShapePurple}{HTML}{7D922A}
\definecolor{ShapePurpleFill}{HTML}{E9F1CB}
\definecolor{InnerStroke}{HTML}{B4B2A9}
\definecolor{InnerFill}{HTML}{F1EFE8}

\usepackage{eso-pic}
\usepackage{xcolor}

\AddToShipoutPictureBG*{%
  \AtPageUpperLeft{%
    \parbox[t][2cm][c]{\paperwidth}{%
      \centering\small\color{gray}
      This is the author's accepted version of a paper accepted for publication\\
      at the International Semantic Web Conference (ISWC) 2026.\\
      The final Version of Record will be published by Springer.
    }%
  }%
}

\begin{document}
\title{NL2SHACL-Bench: A Benchmark Suite for Natural Language to SHACL Translation}
\titlerunning{NL2SHACL-Bench: A Benchmark Suite for NL to SHACL Translation}
%
\author{Yuchen Zhou\inst{1}\textsuperscript{(\Letter)}\orcidlink{0009-0003-0305-9622} \and
Niels Bobet\inst{1}\orcidlink{0009-0009-2738-1847} \and
Maribel Acosta\inst{1}\textsuperscript{(\Letter)}\orcidlink{0000-0002-1209-2868}}
\authorrunning{Y. Zhou et al.}
%
\institute{
Technical University of Munich, Munich, Germany\\
\email{\{yuchen.zhou, niels.bobet, maribel.acosta\}@tum.de}}
\maketitle              
\begin{abstract}
Authoring SHACL shapes to validate the conformance of RDF knowledge graphs (KGs) requires technical expertise that most domain experts lack.
Translating natural language requirements into SHACL (NL2SHACL) would lower this barrier. 
However, there is no dedicated benchmark for this task, and evaluating generated shapes requires methods beyond string comparison, as semantically equivalent shapes can differ in serialisation and structure. 
In this paper, we present NL2SHACL-Bench, a benchmark suite for natural language to SHACL translation.  
Using NL2SHACL-Bench, we evaluate four state-of-the-art large language models (LLMs) for this task. 
Our results show that current LLMs are highly capable of generating syntactically valid SHACL, but still struggle to produce semantically equivalent constraints for complex logical and structural patterns. This indicates that NL2SHACL-Bench provides a meaningful basis for measuring advances in the NL2SHACL state of the art. 
\\
\textbf{Resource Type:} Benchmark\\
\textbf{Source Code:} \href{https://de-tum.github.io/NL2SHACL-Bench}{https://de-tum.github.io/NL2SHACL-Bench}\\
\textbf{License:} MIT license\\
\textbf{Dataset:} \href{https://zenodo.org/records/20082565}{https://zenodo.org/records/20082565}, CC-BY-SA 4.0\\

\end{abstract}
\section{Introduction}
\label{sec:introduction}

The Shapes Constraint Language (SHACL)~\cite{shacl} is a W3C recommendation for validating RDF knowledge graphs (KGs). 
It is widely used to ensure data quality and consistency in KG-based systems, supporting tasks such as compliance checking and metadata validation~\cite{rabbani2022shacl}. 
However, creating SHACL shapes for validating real-world KGs requires expertise in both the application domain and semantic web technologies, which most domain experts lack. 

In practice, constraints are often expressed in natural language by domain experts and then manually translated into SHACL~\cite{luthfi2022sock}. 
This process is time-consuming and error-prone, motivating the need for automated solutions. 
Translating natural language requirements into SHACL (NL2SHACL), therefore, offers a promising way to lower this barrier.

Despite this motivation, NL2SHACL has not yet been systematically studied as a dedicated task. 
Existing SHACL resources are typically mined from structured data~\cite{cimmino2020astrea,fernandez-alvarez2022automatic,rabbani2023extraction} or designed for specific domains~\cite{schaffenrath2020benchmark}, and therefore do not provide aligned natural language descriptions. 
Moreover, evaluating generated SHACL shapes is non-trivial, since semantically equivalent shapes may differ substantially in serialization and graph structure, making string-based comparison insufficient. 
As a result, there is currently no dedicated benchmark, standardized dataset, or evaluation methodology for systematic NL2SHACL evaluation.


To address these challenges, we present \textbf{NL2SHACL-Bench}, a benchmark suite for natural language to SHACL translation. 
Our suite aims at providing datasets and metrics to assess the performance of NL2SHACL solutions. 
To this end, we collect cross-domain and domain-specific openly available SHACL shapes and create a dataset that includes natural language specifications or descriptions of these shapes and ontology annotations. 
To this end, NL2SHACL-Bench includes a software framework for creating such datasets. 
Lastly, we define and implement a set of metrics capturing syntactic, structural, and semantic equivalence. 
In summary, our suite comprises: (i) NL2SHACL-Framework, an extensible pipeline for dataset construction and evaluation; (ii) NL2SHACL-Dataset, currently covering six datasets across five domain-specific settings and one general domain (DBpedia) with 240 manually verified NL–SHACL pairs annotated with ontology information; and (iii) NL2SHACL-Metrics, defining seven metrics across syntactic, structural, and semantic dimensions tailored to assess the quality of translated shapes.
The contributions of this work are as follows:

\begin{itemize}
    \item We introduce \textbf{NL2SHACL-Bench}, the first benchmark suite for NL-to-SHACL translation, comprising a reusable dataset, framework, and evaluation metrics.
    \item We evaluate several state-of-the-art LLMs to demonstrate the usefulness of the benchmark for systematic NL2SHACL evaluation. 
\end{itemize}

The remainder of this paper is organized as follows. Section~\ref{sec:related-work} reviews related work. Section~\ref{sec:preliminaries} introduces the foundations of SHACL and formally defines the NL2SHACL translation task. Section~\ref{sec:nl2shacl-bench} presents NL2SHACL-Bench in detail, describing the framework, dataset, and evaluation metrics. Section~\ref{sec:exp} reports our experimental evaluation of four state-of-the-art LLMs on the benchmark. Finally, Section~\ref{sec:conclusion} concludes the paper and outlines directions for future work.


\section{Related Work}
\label{sec:related-work}

\noindent\textit{Automatic SHACL Generation.}
Authoring SHACL shapes manually is a complex and time-consuming task, which has motivated a range of approaches for automatic shape generation. 
Existing methods primarily derive SHACL shapes from structured artefacts such as ontologies or RDF data.
Astrea~\cite{cimmino2020astrea} generates SHACL shapes from OWL ontologies by mapping ontology axioms to SHACL constraint patterns, using a knowledge graph of mappings to cover both value and model restrictions. 
sheXer~\cite{fernandez-alvarez2022automatic} mines RDF graph topology to infer constraints, supporting both SHACL and ShEx generation. 
QSE~\cite{rabbani2023extraction} targets scalable extraction from very large knowledge graphs, using statistical measures such as support and confidence to filter unreliable constraints. 
SCOOP~\cite{duan2024scoop} extracts constraints from schema definitions, RDF data, and query logs within a single pipeline, while XSD2SHACL~\cite{duan2023xsd2shacl} derives shapes from XML Schema rather than RDF or ontology axioms.
Despite these advances, automatically generated shapes exhibit several limitations. 
In particular, data-driven extraction can introduce spurious constraints due to noisy or incorrect data. 
Besides, these approaches typically generate a large number of candidate shapes, making manual validation and selection costly in practice~\cite{rabbani2022shacl}.
In practice, SHACL constraints are often derived from requirements expressed in natural language by domain experts, which are then translated into formal shapes by knowledge engineers~\cite{luthfi2022sock,makelburg2024automation}, motivating the need for methods that can automate this process.
To the best of our knowledge, this setting has not been systematically studied, which forms the focus of this work.


\noindent\textit{SHACL Benchmarks and Datasets.}
Existing SHACL resources are scattered across public repositories and curated collections. The \textit{Shapes of You} index aggregates SHACL and ShEx shapes from public Git repositories, but many files are incomplete, inaccessible, or lack accompanying ontologies, and numerous repositories contain only a small number of shapes. As a result, these resources are not well-suited for systematic benchmarking.
Several works have proposed datasets involving SHACL shapes~\cite{makelburg2024automation,rabbani2023extraction,schaffenrath2020benchmark}. 
Schaffenrath et al.~\cite{schaffenrath2020benchmark} construct a benchmark with 58 manually designed shapes for validating a tourism knowledge graph. 
Mäkelburg et al.~\cite{makelburg2024automation} provide expert-defined constraints for an electronic invoicing KG.
Rabbani et al.~\cite{rabbani2023extraction} introduce a dataset with SHACL shapes extracted automatically from cross-domain KGs. 
Overall, existing resources either lack natural language annotations, are domain-specific, or are not designed for generation tasks. To date, \emph{no benchmark pairs natural language constraint descriptions with gold-standard SHACL shapes}, nor has a dedicated evaluation methodology been proposed for NL-to-SHACL. This gap motivates the benchmark introduced in this work.

\noindent\textit{LLMs for Shapes Management.}
Zhang et al.~\cite{zhang2025schema} investigate the use of LLMs for generating ShEx schemas for Wikidata and Yago. Their approach combines local entity information with global graph context to produce syntactically valid and semantically meaningful ShEx expressions. However, it focuses on ShEx and on schema extraction from structured knowledge graphs, rather than generation from natural language descriptions.
Westermann et al.~\cite{westermann2025automated} explore the use of LLMs to translate textual constraints into SHACL shapes within an industrial validation pipeline, effectively addressing the same NL-to-SHACL task. Their results show that LLMs can produce near-complete SHACL constraints with limited post-editing. 
However, their evaluation relies on manual expert inspection over a proprietary, single-domain dataset, without standardized metrics or a released benchmark.
Despite these advances, the capability of LLMs to generate SHACL shapes from natural language descriptions has not been systematically studied, and no dedicated benchmark exists for evaluating this task.


\section{Preliminaries and Problem Formulation}
\label{sec:preliminaries}


\begin{wrapfigure}[21]{r}{0.55\textwidth}
\vspace{-6mm}
\resizebox{\linewidth}{!}{%
%


\begin{tikzpicture}[
    font=\small\sffamily,
    nodeshape/.style={
        rectangle, rounded corners=10pt,
        line width=0.6pt, inner sep=10pt,
    },
    propshape/.style={
        rectangle, rounded corners=4pt,
        draw=InnerStroke, line width=0.4pt, fill=white,
        inner sep=6pt, text width=5.5cm, align=left,
    },
    targetbox/.style={
        rectangle, rounded corners=3pt,
        draw=InnerStroke, line width=0.4pt, fill=InnerFill,
        inner sep=5pt, minimum width=5.5cm, align=center,
    },
    shapehead/.style={align=center, text width=5.5cm},
    annotation/.style={
        font=\footnotesize\itshape\sffamily, text=gray!50!black,
    },
    ref/.style={
        -{Stealth[length=2.5mm]},
        draw=ShapePurple, line width=0.7pt,
    },
    leader/.style={
        -{Stealth[length=1.6mm]},
        draw=gray!60, densely dashed, line width=0.4pt,
    },
]

\node[shapehead] (pTitle)
    {ex:PersonShape 
     \textit{\footnotesize a sh:NodeShape}};
\node[targetbox, below=5pt of pTitle] (pTarget)
    {sh:targetClass \textbf{ex:Person}};
\node[propshape, below=8pt of pTarget] (ps1)
    {sh:path \textbf{ex:status}\\
     sh:in (``active'', ``inactive'')};
\node[propshape, below=6pt of ps1] (ps2)
    {sh:path \textbf{ex:address}\\
     sh:node ex:AddressShape};

\begin{scope}[on background layer]
    \node[nodeshape, draw=ShapeTeal, fill=ShapeTealFill,
          fit=(pTitle)(pTarget)(ps1)(ps2)] (personShape) {};
\end{scope}

\node[shapehead, below=1cm of personShape] (aTitle)
    {ex:AddressShape 
     \textit{\footnotesize a sh:NodeShape}};
\node[targetbox, below=5pt of aTitle] (aTarget)
    {sh:targetClass \textbf{ex:Address}};
\node[propshape, below=8pt of aTarget] (ps3)
    {sh:path \textbf{ex:postalCode}\\
      sh:or (\\
     \quad{}[\,sh:pattern \texttt{"\^{}[0-9]\{5\}\$"}\,],\\
     \quad{}[\,sh:pattern \texttt{"\^{}[A-Z]\{2\}[0-9]\{4\}\$"}\,]\\
     )};

\begin{scope}[on background layer]
    \node[nodeshape, draw=ShapePurple, fill=ShapePurpleFill,
          fit=(aTitle)(aTarget)(ps3)] (addressShape) {};
\end{scope}

\draw[ref] (ps2.south) --
    node[pos=0.5, inner sep=4pt,
         font=\footnotesize\sffamily] {\textit{reference}}
    (addressShape.north);

\node[annotation, right=18pt of ps1] (ann1) {sh:PropertyShape};
\draw[leader] (ann1.west) -- (ps1.east);

\node[annotation, right=18pt of ps2] (ann2) {sh:PropertyShape};
\draw[leader] (ann2.west) -- (ps2.east);

\node[annotation, right=18pt of ps3] (ann3) {sh:PropertyShape};
\draw[leader] (ann3.west) -- (ps3.east);

\end{tikzpicture}
}
\caption{Example: schematic representation of a SHACL shapes graph.}
    \label{fig:shacl}
\end{wrapfigure}
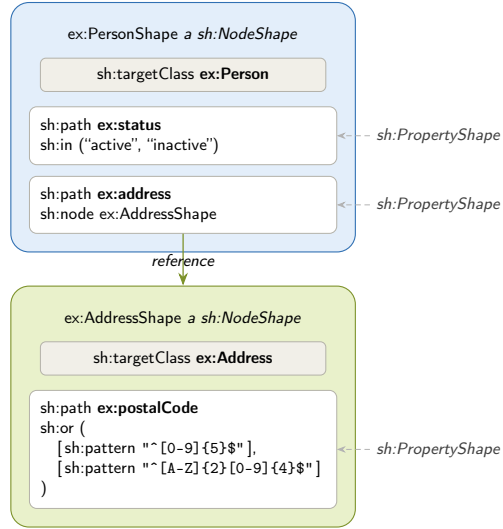

SHACL operates over two graphs: data graph and shapes graph.  
The \textit{data graph} corresponds to the RDF graph $\mathcal{G}$ to be validated. We assume $\mathcal{G}$ is described with an ontology $O$. 
In general, a shapes graph is an RDF graph containing zero or more shapes, where both node shapes and property shapes may independently declare targets and trigger validation~\cite{shacl}. The definitions below focus on \textit{node-shape-rooted} shapes graphs, where every property shape is either inlined within a node shape via \texttt{sh:property} or reachable from one. Under this scope, a shapes graph $\mathcal{S}$ is a set of node shapes $\{s_1, \cdots, s_n\}$.

A \textit{node shape} $s_i$ declares constraints that apply to focus nodes, which are selected via a \textit{target}, e.g., \texttt{sh:targetClass}, which selects all instances of a given class in $O$.
Other targets include \texttt{sh:targetSubjectsOf}, and \texttt{sh:targetObjectsOf}, or \texttt{sh:targetNode}.
Within a node shape, \textit{property shapes} (introduced via \texttt{sh:property}) describe constraints on values reachable through a specific path, using predicates defined in $O$, declared with \texttt{sh:path}. 
The actual restrictions come from \textit{constraint components} such as \texttt{sh:and}, \texttt{sh:or}, \texttt{sh:in}, \texttt{sh:minCount}, \texttt{sh:datatype}, \texttt{sh:pattern}, \texttt{sh:class}, or \texttt{sh:nodeKind}, each parameterized by values the shape author supplies. 
A node shape $s_i$ can contain references to another node shape $s_j$ (allowing $i=j$). 
Figure~\ref{fig:shacl} illustrates these concepts. 
A shapes graphs $\mathcal{S}$ can be serialized as an RDF graph, and we denote $|\mathcal{S}|$ as its number of RDF triples.

The validation of a $\mathcal{S}$ over a $\mathcal{G}$ produces a report with \textit{conformance} results.  
$\mathcal{G} \models \mathcal{S}$ denotes that the data graph conforms to the shapes graph~\cite{ahmetaj2025common}. Two shapes graphs $\mathcal{S}$ and $\mathcal{S}'$ are \emph{equivalent}, written
$\mathcal{S} \equiv \mathcal{S}'$, iff
$\forall \mathcal{G}: \ \mathcal{G} \models \mathcal{S} \ \Leftrightarrow\ \mathcal{G} \models \mathcal{S}'$.

Following a task formulation similar to \cite{ahmetaj2025common}, we define the NL2SHACL translation task as follows.

\begin{definition}[NL2SHACL Translation]
\label{def:nl2shacl}
Given a natural language specification $T$ describing constraints over a domain, and an ontology $O$ providing the vocabulary (classes and properties) referenced by $T$. 
The task is to produce a shapes graph $\mathcal{S}$ over $O$ s.t. $\mathcal{S}$ encodes the constraints expressed in $T$. 
A translator is a mapping $f: (T, O) \mapsto \mathcal{S}$. We say $f$ is \textit{correct} on $(T, O)$ iff $\mathcal{S} = f(T, O)$ is equivalent to a reference shapes graph $\mathcal{S}^*$ that captures the semantics of $T$, i.e.,

$$\forall \mathcal{G} \text{ over } O:\quad \mathcal{G} \models f(T, O) \;\Leftrightarrow\; \mathcal{G} \models \mathcal{S}^*.$$
\label{def:problem}
\end{definition}

\section{NL2SHACL-Bench}
\label{sec:nl2shacl-bench}

To address the lack of dedicated resources for natural language to SHACL translation and the limitations of existing evaluation approaches, we present \textbf{NL2SHACL-Bench}, a benchmark suite for the NL2SHACL task. 
This benchmark is designed to provide a standardized basis for evaluating model performance on translating natural language requirements into SHACL constraints. 
NL2SHACL-Bench consists of three main components: 
an extensible framework for dataset construction and evaluation (\S\ref{sec:framework}); 
a semi-automatically curated dataset covering multiple domains (\S\ref{sec:nl2shacl-dataset});
and a set of tailored evaluation metrics (\S\ref{sec:nl2shacl-metrics}).
In the following, we introduce each component in detail.

\subsection{NL2SHACL-Framework}
\label{sec:framework}

\begin{figure}[t!]
    \centering
    \includegraphics[width=\linewidth]{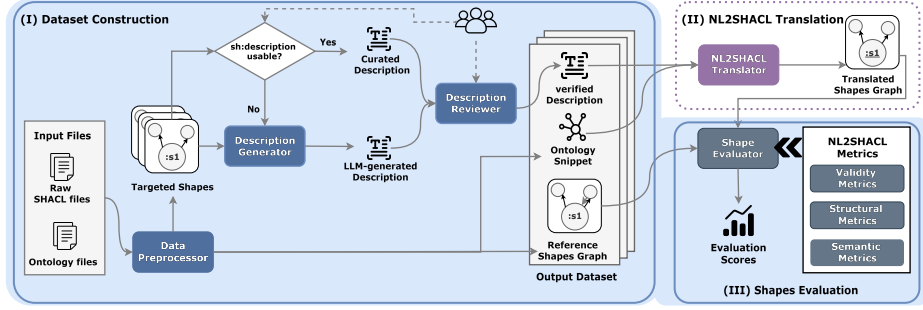}
    \caption{Overview of the NL2SHACL framework. The design is modular and standardized, allowing flexible benchmarking of NL2SHACL systems.}
    \label{fig:framework}
\end{figure}

We propose an extensible framework that supports the end-to-end benchmarking of NL2SHACL systems. 
As shown in Figure~\ref{fig:framework}, our framework consists of three modules. 
(i)~dataset construction, (ii)~NL2SHACL translation, and (iii)~shapes evaluation. 
Our framework is implemented in Python and is publicly available.

NL2SHACL-Framework further exhibits several desirable properties. 
First, it is \textbf{modularized}: each module can be used independently, allowing users to focus on dataset construction, model evaluation, or the full pipeline as needed. 
Second, it is \textbf{extensible}: users can incorporate custom input files to generate new datasets, replace the translation module with arbitrary models, and extend the evaluation module with additional metrics.
Third, it is also \textbf{reproducible}, as it defines a consistent pipeline for dataset construction and evaluation. 
Overall, the framework is designed to standardize the benchmarking pipeline for NL2SHACL. 
In the following, we describe each module in detail.

\subsubsection{Dataset Construction Module}
This module is a semi-automated pipeline for constructing a paired dataset for the NL2SHACL task. 
In practice, as discussed in Section~\ref{sec:related-work}, shapes graphs lack natural language descriptions. 
Therefore, this module creates these descriptions from a reference shapes graph and an ontology. 
To transform these inputs into structured NL--SHACL pairs, the module comprises three components: data pre-processing, description construction, and description reviewing. 


\paragraph{Component 1: Data Pre-processor.}
For a given shapes graph $\mathcal{S}^*$, we extract individual \textit{shape fragments}. 
Each shape fragment (or just fragment) corresponds to a node shape together with all the property shapes and other node shapes it references; this is extracted from the shape-based constraint components (e.g., \texttt{sh:node}, \texttt{sh:property}).\footnote{In our current implementation, we exclude SPARQL-based constraints as NL2SPARQL is a separate research problem that requires dedicated methods.}  
In this way, each fragment represents a complete constraint subgraph without requiring external shape definitions.
With this definition, a fragment is also a shapes graph, which we denote $\mathcal{S}_i^*$. 
An example of a fragment with two node shapes is shown in Figure~\ref{fig:shacl}.
%
Then, the component performs a structural completeness check to ensure that each extracted fragment yields a valid SHACL shapes graph. This check verifies that every \texttt{sh:property} block carries a \texttt{sh:path} declaration, and that the fragment contains at least one \texttt{sh:NodeShape}; fragments failing either condition are removed.
Lastly, for each fragment, this component extracts ontology term URIs from all SHACL predicates that can reference domain vocabulary. 
This produces an \textit{ontology snippet} $O_i$: a compact, prompt-ready extract of exactly the vocabulary referenced by the fragment.  
Fragments with fully unresolved domain terms are discarded.

\paragraph{Component 2: Description Constructor.}
This component generates a natural language description $T_i$ for each shape fragment $\mathcal{S}_i^*$.
First, we check whether a fragment already contains a \texttt{sh:description}, which may have been provided by a human during the creation of the SHACL shape. 
If present, it is manually reviewed to assess its usability. 
Descriptions that are deemed usable are curated and directly retained.
If the description is missing or not directly usable (e.g., incomplete or lacking sufficient detail to reflect the underlying constraints), we generate a description draft using an LLM.
Specifically, we prompt the model with the reference SHACL shapes in $\mathcal{S}_i^*$ and a carefully designed instruction that guides the translation of SHACL constraints and ontology terms into fluent, domain-appropriate natural language. 
The model is assigned the role of a domain expert who is not familiar with RDF or SHACL, and is asked to express the underlying data constraints in natural language as they would in practice, using a clear and sufficiently detailed description. 
This design aims to simulate how human domain experts articulate validation requirements, making the generated descriptions more realistic and closer to real-world usage. 
The model is instructed to produce a single coherent paragraph per fragment, following a consistent register and format within each subset. 
The output of this component is a description draft for each fragment, regardless of whether it originates from existing annotations or LLM generation.

\paragraph{Component 3: Description Reviewer.}
To ensure the quality of LLM-generated descriptions $T_i$ and mitigate issues such as hallucinations, we introduce a human validation component. To support this process, we develop a graphical user interface (GUI) to assist annotators in reviewing and editing descriptions.
We conduct a two-round review process involving two annotators with experience in SHACL.
In the first round, the annotators independently review each description alongside its corresponding shape fragment, identifying issues across six categories: incorrectly described constraints, missing or redundant information, overly technical phrasing, unnatural language, exposure of SHACL-specific terminology, and content hallucinated by the model beyond what the shape encodes. The latter two categories specifically target LLM-friendly phrasing patterns that could otherwise bias downstream evaluation.
In the second round, the annotators discuss the flagged cases and agree on the necessary revisions. Through this process, the final description $T^*_i$ are refined to faithfully and completely reflect the constraints encoded in the corresponding SHACL shapes.

\begin{figure}[t!]
    \centering
    \includegraphics[width=\linewidth]{figures/example_datapoint.pdf}
    \caption{An example record from the Invoice dataset, showing a reference shapes graph $\mathcal{S}^{*}_i$, the natural 
    language description $T^*_i$, and the relevant ontology snippet $O_i$.}
    \label{fig:data_example}
\end{figure}

\paragraph{Output Dataset.}
The output of this module is a dataset with \textit{records} in a unified format. 
Each record is a triple $(\mathcal{S}^*_i, T^*_i, O_i)$ of a reference shapes graph $\mathcal{S}_i^*$, a natural language description $T^*_i$, and an ontology snippet $O_i$ for the referenced terms.
In the remainder of this paper, we use \textit{record} and \textit{data point} interchangeably to refer to this triple.
Figure~\ref{fig:data_example} shows an example of a record created by our framework.

\subsubsection{NL2SHACL Translator Module}
This is an off-the-shelf module that provides the translation mapping $f$ (Definition~\ref{def:problem}). 
Given a natural language description $T_i^*$ and the corresponding ontology snippet $O_i$, $f$ produces a translated shapes graph $\mathcal{S}_i=f(T_i^*,O_i)$.
The module is model-agnostic. 
Users can plug in arbitrary translation systems, including LLM-based or rule-based approaches.
In our evaluation, we instantiate this module using a prompting-based LLM translator, which is described in Section~\ref{sec:exp}.

\subsubsection{Shapes Evaluation Module}
This module evaluates the quality of the translated shapes $\mathcal{S}_i$. 
The evaluation is performed using a set of predefined metrics, which produce quantitative scores reflecting different aspects of quality. 
The module is extensible and allows additional metrics to be incorporated.
The metrics used in this work are described in Section~\ref{sec:nl2shacl-metrics}.

\subsection{NL2SHACL-Dataset}
\label{sec:nl2shacl-dataset}

\subsubsection{Dataset Overview}
The NL2SHACL dataset comprises multiple sub-datasets covering both cross-domain (DBpedia) and domain-specific settings, including business invoices, chemical data, healthcare information management, public procurement, and open government data.
This dataset has been created with the Dataset Construction Module; therefore, each data point follows a unified structure consisting of a natural language description $T_i^*$, a reference shapes graph $\mathcal{S}_i^*$, and an ontology snippet $O_i$ capturing the semantics of the terms referenced in the shapes graph.
%
In total, the dataset contains 240 data points across 6 sub-datasets. Table~\ref{tab:dataset_details} summarizes the statistics of each subset.
The dataset covers 21 of the 33 SHACL Core constraint components (excluding SHACL-SPARQL). The absent components reflect the characteristics of the selected source repositories rather than limitations of the framework, which can process any of these components when present in source data.\footnote{A full per-subset coverage breakdown, including the absent components, is available at \url{https://github.com/DE-TUM/NL2SHACL-Dataset\#constraint-component-coverage}.}

\begin{table}[t!]
\centering
\caption{NL2SHACL benchmark statistics and filtering summary.}
\label{tab:dataset_details}
\begin{threeparttable}
{\scriptsize
\begin{tabular}{l rrrrr | rr}
\toprule
& \multicolumn{5}{c|}{\textbf{\scriptsize Dataset Statistics}} & \multicolumn{2}{c}{\textbf{\scriptsize Shape Filtering}} \\
\cmidrule(lr){2-6} \cmidrule(lr){7-8}
\textbf{Subset}
  & \thead{\textbf{\scriptsize \# Data}\\\textbf{\scriptsize Records}}
  & \thead{\textbf{\scriptsize \# Node}\\\textbf{\scriptsize Shapes}}
  & \thead{\textbf{\scriptsize \# Prop.}\\\textbf{\scriptsize Shapes}}
  & \thead{\textbf{\scriptsize Avg. NL}\\\textbf{\scriptsize Length}}
  & \thead{\textbf{\scriptsize Avg. Shapes}\\\textbf{\scriptsize per Record}}
  & \thead{\textbf{\scriptsize \# Raw}\\\textbf{\scriptsize Records}}
  & \thead{\textbf{\scriptsize \# Filtered}\\\textbf{\scriptsize Records}} \\
\midrule
CHEMROF~\cite{chemkgchemrof}  &  74 &  74 &  441 & 154.5 &  6.96 & 128 & 17 \\
DCAT~\cite{dcatapshacl}       &  20 &  35 &  103 & 116.2 &  6.90 &  21 &  1 \\
ePO~\cite{eprocurementontology} & $^\dagger$50 & 50 & 139 & 116.5 & 3.78 & 378 & 235 \\
Invoice~\cite{edifactval}     &  78 &  78 &  113 &  75.1 &  2.45 &  84 &  6 \\
SNIK~\cite{snikproject}       &   8 &  11 &   49 &  52.0 &  7.50 &  27 &  0 \\
DBpedia~\cite{auer2007dbpedia}               &  10 &  10 &   11 &   9.5 &  2.10 &  -- & -- \\
\midrule
\textbf{Overall} & \textbf{240} & \textbf{258} & \textbf{856} & \textbf{108.1} & \textbf{4.64} & -- & -- \\
\bottomrule
\end{tabular}
}
\begin{tablenotes}
\scriptsize
\item[$\dagger$] Sampled from 143 records.
\end{tablenotes}
\end{threeparttable}
\end{table}

\subsubsection{Data Sources and Construction}
The dataset of reference shapes $\mathcal{S}^*$ is constructed from multiple data sources. 
Four subsets are selected from the \texttt{Shapes of You} index. 
In addition, the Invoice subset is derived from our previous work, and the DBpedia subset is curated by the authors using the DBpedia ontology.

We select the four repositories from the index\footnote{From over 120 SHACL repositories initially collected, only a small number were suitable for inclusion due to issues such as inaccessibility, missing ontologies, or insufficient numbers of usable shapes.} due to the following desirable properties: (1) each represents a well-defined application domain; (2) an associated ontology is available to support ontology-grounded description and evaluation; and (3) each contains a sufficient number of shapes (more than 10\footnote{Determined empirically: repositories below this size do not represent a domain's constraint patterns meaningfully, and annotation cost is fixed regardless of repository size.}).
The selected shapes encode meaningful constraints that are neither trivially shallow nor excessively complex.
Table~\ref{tab:dataset_details} summarizes the filtering process and the number of remaining records for each subset.


\subsubsection{Sub-dataset Description}
Below, we briefly describe each subset.

\smallskip
\noindent\textbf{CHEMROF.} The CHEMROF data source~\cite{chemkgchemrof} models chemical entities and their relationships, ranging from atoms to complex mixtures. The shapes specify constraints such as closed-world restrictions, mandatory identifiers, enumerated values, and hierarchical classifications.

\smallskip
\noindent\textbf{DCAT.} The DCAT data source~\cite{dcatapshacl} is based on the DCAT-AP vocabulary for describing datasets and metadata in European data portals. The shapes define constraints on metadata properties such as license, language, and publisher, often requiring values to link to controlled vocabularies. Notably, this subset includes nested constraints via \texttt{sh:node}, introducing multi-level validation dependencies.

\smallskip
\noindent\textbf{ePO.} The eProcurement Ontology (ePO)~\cite{eprocurementontology} models public procurement processes in the European Union, covering multiple stages such as ordering, invoicing, and payment. The shapes specify constraints on entities and relationships across these stages, resulting in a wide variation in complexity. Additional filtering and sampling are applied to ensure balanced and tractable descriptions.

\smallskip
\noindent\textbf{Invoice.} The Invoice data source~\cite{edifactval} is designed for validating invoicing data conforming to the EDIFACT standard\footnote{Electronic Data Interchange for Administration, Commerce and Transport}. The shapes specify constraints such as required fields, data formats, and identifier structures. Notably, several shapes employ \texttt{sh:or} and \texttt{sh:not} to express alternative and negation constraints.

\smallskip
\noindent\textbf{SNIK.} The SNIK data source~\cite{snikproject} models information management in healthcare systems using a meta-level ontology. The shapes primarily define semantic constraints through class assignments and target declarations (e.g., \texttt{sh:class}, \texttt{sh:targetClass}), with relatively few property-level restrictions. This results in structurally simple but semantically expressive constraints.

\smallskip
\noindent\textbf{DBpedia.} The DBpedia~\cite{auer2007dbpedia} subset is curated using the DBpedia ontology and focuses on a wide range of constraint types. The shapes include not only existence, cardinality, and datatype constraints, but also logical and conditional rules. Unlike other subsets, the natural language descriptions are derived from and reviewed based on existing \texttt{sh:description} annotations.
 

\subsection{NL2SHACL-Metrics}
\label{sec:nl2shacl-metrics}

In NL2SHACL, Evaluating  the correctness of a translated shapes graph $\mathcal{S}_i=f(T_i^*,O_i)$ w.r.t. the reference $\mathcal{S}^*$ requires going beyond surface-level comparison, as syntactically different shapes may still be semantically equivalent.  
We therefore design a set of metrics that assess generated shapes from three complementary perspectives: validity, structure, and semantics.
Together, these metrics provide a comprehensive and fine-grained evaluation of NL2SHACL systems.

\subsubsection{Validity Metrics}
We define three validity metrics capturing different levels of correctness, organized as a layered pipeline where each stage is only reached if the previous one passes:

\begin{enumerate}
    \item \textbf{RDF Parsing Validity:} Checks whether $\mathcal{S}_i$ is a valid RDF graph via standard parsing. Outputs that fail this check are recorded as a \textit{Parsing Error}.
    \item \textbf{SHACL Specification Validity:} Validates $\mathcal{S}_i$ against the W3C SHACL-SHACL meta-shapes, applied only to outputs that pass RDF parsing. Outputs that fail this check are recorded as a \textit{Specification Error}.
    \item \textbf{SHACL Vocabulary Validity:} Checks all \texttt{sh:}-prefixed predicates in $\mathcal{S}_i$ against the set of standard SHACL constraint components, applied to outputs that pass both previous checks. This layer targets a failure mode specific to LLM translators: hallucinated non-standard vocabulary (e.g., \texttt{sh:if}, \texttt{sh:then}, \texttt{sh:condition}). Shapes that fail this check are recorded as a \textit{Vocabulary Error}.
\end{enumerate}

For each data point, a binary result is assigned at each stage. Each validity rate (VR) is computed as the proportion of outputs that pass the corresponding check among those that reach that stage. Let $N$ be the total number of data points, $N_\text{rdf}$ the number that pass RDF parsing, and $N_\text{spec}$ the number that additionally pass the specification check. Let $P_i, S_i, V_i \in \{0,1\}$ denote whether $\mathcal{S}_i$ satisfies RDF parsing, SHACL specification, and vocabulary validity, respectively. The validity rates are defined as follows:
\[
\text{RDF-VR} = \frac{1}{N}\sum_{i=1}^{N} P_i, \ 
\text{Spec-VR} = \frac{1}{N_\text{rdf}}\sum_{i=1}^{N} P_i S_i, \ 
\text{Vocab-VR} = \frac{1}{N_\text{spec}}\sum_{i=1}^{N} P_i S_i V_i
\]

For the cases $N_\text{rdf}=0$ or $N_\text{spec}=0$, the metrics $\text{Spec-VR}$ and $\text{Vocab-VR}$ are considered undefined, respectively. 

\subsubsection{Structural Metrics}

We define two structural metrics to measure graph-level structural similarity between the generated shape $\mathcal{S}_i$ and the reference shape $\mathcal{S}^{*}_i$. To enable comparison, both graphs are canonicalized to handle blank nodes, and their triple sets are compared. 
The defined metrics are: 
\begin{itemize}
    \item \textbf{Exact Match:} A binary metric indicating whether the generated and reference shapes are structurally identical after canonicalization, i.e, $\mathbf{1}_{[\mathcal{S}_i=\mathcal{S}^*_i]}$.
    \item \textbf{Partial Match:} A continuous metric measuring the degree of overlap between the two shapes graphs based on triple-level similarity, computed using precision $P=\frac{|\mathcal{S}_i \cap \mathcal{S}^*_i|}{|\mathcal{S}_i|}$, recall $R=\frac{|\mathcal{S}_i \cap \mathcal{S}^*_i|}{|\mathcal{S}^*_i|}$, and $F1=\frac{2PR}{P+R}$. 
\end{itemize}

Exact matching provides a strict criterion, while partial matching offers a finer-grained similarity signal. However, both metrics are inherently structural and do not capture whether two shapes enforce the same constraints, which motivates the semantic metrics introduced next. 

At the dataset level, we report the \textbf{Exact Match Rate} (EMR), defined as the proportion of records with an exact match. We also report the \textbf{Partial Match Score} (PMS) as the macro F1-score across all records.

\subsubsection{Semantic Metrics}
We define semantic metrics to assess whether the translated shapes $\mathcal{S}_i$ and the reference shapes $\mathcal{S}^*_i$ enforce the same constraints. We approach this from two complementary angles: a sampling-based comparison of validation outcomes on RDF data graphs, and a logic-based check of shape containment.

\smallskip\noindent\emph{Approximate Semantic Equivalence.}
For each record, we generate a synthetic RDF data graph $\mathcal{G}_i$ using RDFGraphGen~\cite{jovanovik2025rdfgraphgen}, with the reference shape serving as the schema. This ensures that $\mathcal{G}_i$ contains nodes that are relevant for the validation. Then, we assess the validation conformance, i.e., $\mathcal{G}_i \models \mathcal{S}^*_i$ and $\mathcal{G}_i \models \mathcal{S}_i$. For non-conformance cases, we extract the non-conforming nodes from $\mathcal{G}_i$ and compare them as sets. A binary score is assigned: two shapes are considered equivalent if and only if the sets of violating nodes are identical. As this judgement is made on a sampled data graph rather than on all graphs over $O$, it approximates the equivalence of Definition~\ref{def:nl2shacl}. At the dataset level, we report the \textbf{Approximate Semantic Equivalence Rate (ASER)}, defined as the fraction of records judged equivalent among those for which a data graph could be generated and validated.

\smallskip\noindent\emph{Logical Equivalence.}
To complement the sampling-based judgement, we additionally decide equivalence at the logical level. We say that $\mathcal{S}$ is contained in $\mathcal{S}'$, written $\mathcal{S} \sqsubseteq \mathcal{S}'$, iff $\forall \mathcal{G}: \mathcal{G} \models \mathcal{S} \Rightarrow \mathcal{G} \models \mathcal{S}'$. The biconditional of Definition~\ref{def:nl2shacl} splits into two such checks, so $\mathcal{S}_i \equiv \mathcal{S}^*_i$ holds iff both $\mathcal{S}_i \sqsubseteq \mathcal{S}^*_i$ and $\mathcal{S}^*_i \sqsubseteq \mathcal{S}_i$. We translate both shapes graphs into first-order logic using the SHACL2FOL tool~\cite{pareti2024shacl2fol} and discharge each direction with an automated theorem prover. Checking both directions further reveals the direction of a mismatch, i.e., whether the translated shape is stricter or more permissive than the reference. Since SHACL2FOL supports a fragment of SHACL and the prover operates under a time budget, each record yields one of three outcomes: equivalent, non-equivalent, or undetermined. At the dataset level, we report the \textbf{Logical Equivalence Rate (LER)} as the fraction of equivalent records among those with a determined outcome, together with the \emph{coverage}, i.e., the proportion of evaluated records for which a determined outcome is obtained.

\smallskip
The two metrics are complementary because their failure modes are disjoint. 
ASER inherits the coverage of the generated data graph, which RDFGraphGen produces from the reference shape alone and with limited diversity, so a translated shape with an incorrect target declaration may select no node and be judged equivalent. 
LER is unaffected, since a counterexample graph need not be generated to be proven to exist. 
LER in turn returns undetermined on constraint components outside the supported fragment and on large cardinalities that exceed the time budget~\cite{ahmetaj2025shacl}, where ASER remains available. 

\section{Evaluation} 
\label{sec:exp}

We evaluate the benchmarking capabilities of NL2SHACL-Bench, using the constructed datasets (Section~\ref{sec:nl2shacl-dataset}) and defined metrics (Section~\ref{sec:nl2shacl-metrics}). 
In our experiments, we instantiate the NL2SHACL translator module using LLMs and assess their performance in the NL2SHACL task.
We introduce our evaluation pipeline and experimental results in this section.

\begin{figure}[t!]
    \centering
    \includegraphics[width=\linewidth]{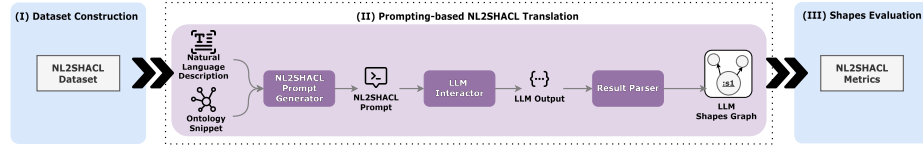}
    \caption{Overview of the evaluation pipeline for the NL2SHACL-Bench.}
    \label{fig:eval-pipeline}
\end{figure}

\begin{figure}[t!]
    \centering
    \includegraphics[width=\linewidth]{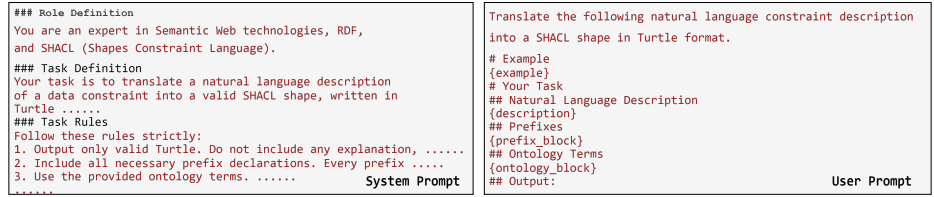}
    \caption{Illustration of the NL2SHACL prompt template used in our evaluation pipeline. The left part shows the \texttt{system prompt}, and the right part shows the \texttt{user prompt}}.
    \label{fig:prompt-template}
\end{figure}

\subsection{Evaluation Set Up}
Figure~\ref{fig:eval-pipeline} shows our evaluation pipeline. 
We use NL2SHACL-Dataset as test data and prompt selected LLMs with pre-defined prompt templates for the task. 
We use the NL2SHACL-Metrics, to assess the shapes generated by LLMs from the given natural language descriptions. 
The details are described as follows.

\smallskip
\noindent
\textbf{Prompt}
We define a prompt template consisting of a static \texttt{system prompt} and dynamically generated \texttt{user prompts}. 
Figure~\ref{fig:prompt-template} illustrates the overall prompt structure.
The system prompt defines the task, role, and generation rules. 
The user prompt has a \textit{one-shot example} together with the test input and output indicator.
Following our task definition, the models are provided with the \texttt{natural language description} $T^*_i$, and relevant \texttt{ontology snippet} $O_i$ of the target shape fragment $\mathcal{S}^*_i$. 
The one-shot example additionally includes the expected SHACL output to demonstrate the desired generation format and support in-context learning, following recent work on LLM-based SHACL generation~\cite{westermann2025automated}. 
All test instances within the same subset share the same example, constructed from the last data item of that subset, which is excluded from evaluation.

\smallskip
\noindent
\textbf{Models}
We select four state-of-the-art LLMs to stress-test our benchmark, covering both proprietary (Gemini 3.1 Pro Preview, Claude Opus 4.7) and open-source (GLM-5.1, Qwen3.5-397B-A17B) providers, chosen based on their strong performance on the Coding Index Score in OpenRouter~\cite{openrouter}.
All models are accessed via the LLM Interactor using the OpenRouter API.


\subsection{Results}

We evaluated a total of 234  records on four LLMs, resulting in 936 translated shape fragments overall.
Reported results are based on post-processed outputs.

\begin{figure}[t]
    \centering
    \includegraphics[width=0.9\linewidth]{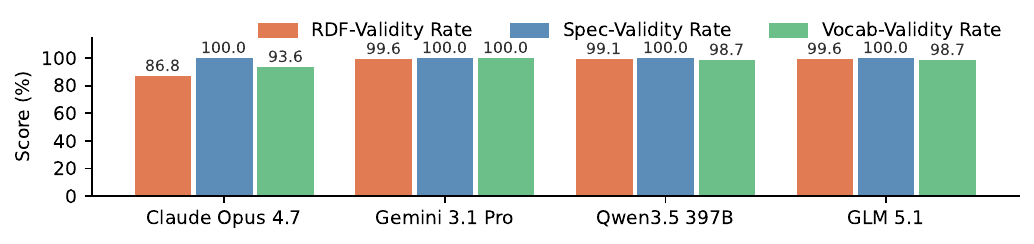}
    \caption{Validity results of the evaluated LLMs on NL2SHACL-Bench.}
    \label{fig:validity-results}
\end{figure}

\subsubsection{Validity Results}
Figure~\ref{fig:validity-results} presents the validity results of the evaluated models across all datasets. 
Overall, all four models achieved very strong validity performance, with Gemini 3.1 Pro obtaining perfect scores on all three metrics and the remaining models producing only a few invalid shapes.
Most RDF parsing failures are low-level formatting issues rather than fundamental misunderstandings of SHACL syntax. 
Claude Opus 4.7 consistently failed due to missing declarations of the \texttt{rdfs:} prefix, whereas Qwen 3.5 and GLM 5.1 produced more isolated errors such as malformed string literals, incorrect escaping, and mismatched bracket structures. 
The latter three models also occasionally hallucinated non-standard components such as \texttt{sh:if}, \texttt{sh:then}, and \texttt{sh:condition} when translating logical constraints expressed through \texttt{sh:or} and \texttt{sh:not}, replacing valid SHACL patterns with invented rule-like constructs.

\vspace{-3mm}
\subsubsection{Structural Results}
\begin{figure*}[t]
    \centering
    \includegraphics[width=\textwidth]{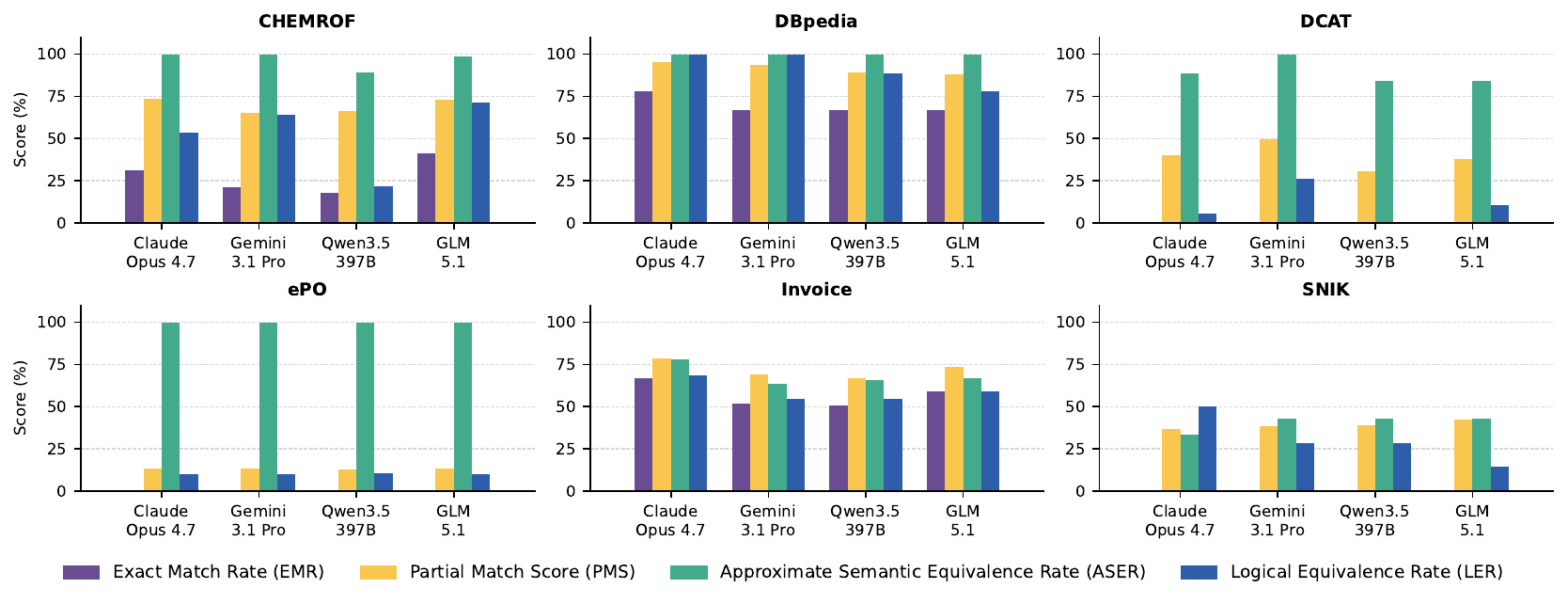}
    \caption{Structural and semantic evaluation results across datasets. Note that both ASER and LER are reported as the proportion of records that were automatically evaluated and judged equivalent. Records that could not be automatically evaluated are excluded from the numerator but retained in the denominator. Consequently, neither rate is binary, and {1 - ASER} or {1 - LER} should not be read as a non-equivalence rate.}
    \label{fig:structural-results}
\end{figure*}

%
Overall, the structural scores vary substantially across datasets as shown in Figure~\ref{fig:structural-results}. 
DBpedia and Invoice achieve comparatively higher EMR and PMS scores, whereas SNIK, DCAT, and ePO show near-zero EMR and consistently low PMS values. 
These differences mainly reflect variations in the structural organization of the reference shapes. 
For SNIK, DCAT, and ePO, the reference shapes often contain multiple interconnected node and property shapes, while the evaluated LLMs typically generate a single node shape with nested constraints. 
As a result, the generated graphs differ substantially from the references at the structural level even when the intended constraints are similar. 
In contrast, DBpedia and Invoice follow more regular and comparatively simple structural patterns, making them easier to reproduce at the graph level.

A clear gap can be observed between EMR, PMS, and ASER across most datasets. 
EMR consistently yields the lowest scores due to its strict exact-matching requirement, while PMS is slightly higher by allowing partial structural overlap. 
At the same time, the comparatively higher ASER scores indicate that structural mismatch does not necessarily imply semantic mismatch. 
Therefore, EMR and PMS should mainly be interpreted as lower-bound indicators of graph-level similarity rather than direct measures of semantic correctness.

\subsubsection{Semantic Results}

\begin{figure*}[t]
    \centering
    \includegraphics[width=0.9\linewidth]{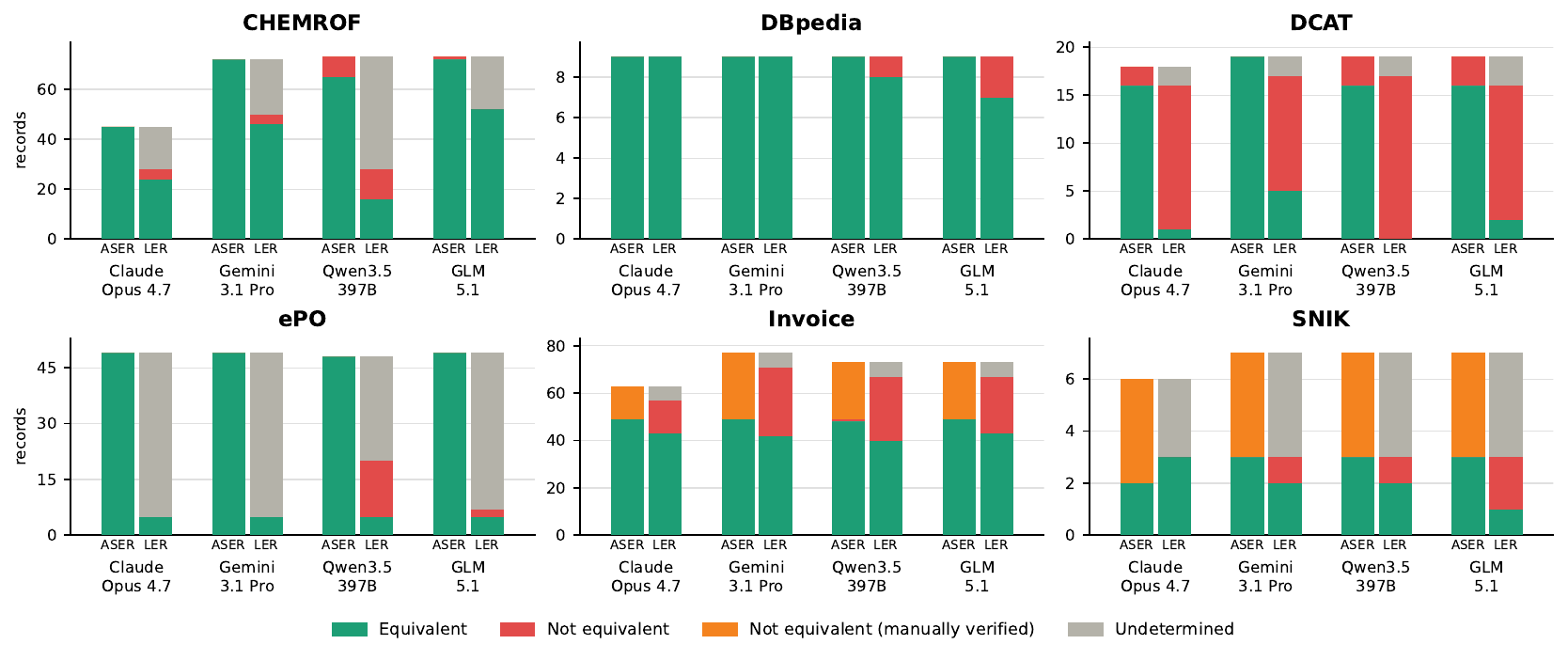}
    \caption{Semantic equivalence results across datasets and models.}
    \label{fig:semantic-results}
\end{figure*}

Figure~\ref{fig:structural-results} reports the aggregate ASER and LER values for each dataset and model. Figure~\ref{fig:semantic-results} breaks each rate down into its underlying categories: equivalent, non-equivalent, non-equivalent confirmed through manual inspection, and undetermined. DBpedia shows the highest agreement between ASER and LER among all datasets, while Invoice, CHEMROF, and SNIK show moderate agreement. ePO and DCAT show the clearest disagreement between the two metrics.

Manual inspection of these disagreements reveals several recurring patterns. First, ASER cannot detect LLM over-specification, while LER can. For example, on DBpedia, Qwen3.5 adds an extra \texttt{sh:minCount 1} constraint not licensed by the natural language description. Second, ASER cannot generate a data graph for complex constraint combinations and requires manual judgement, such as \texttt{sh:or} and \texttt{sh:not} combinations on Invoice, while LER correctly verifies these records as non-equivalent. Third, ASER cannot detect a divergent target declaration written by the model, which results in false positives, such as on DCAT, while LER successfully detects this. Fourth, due to the finite-model-building search underlying LER, it times out on certain constraint combinations, resulting in undetermined outcomes on records such as those on Invoice and a large share of ePO shapes.
Overall, the limitations of the two metrics are disjoint and highly complementary.

Across models, the two metrics support consistent rankings on the datasets where both achieve good coverage. Gemini 3.1 Pro and Claude Opus 4.7 lead on DBpedia and Invoice under both ASER and LER. Qwen3.5 and GLM-5.1 accumulate more non-equivalent or undetermined LER outcomes on CHEMROF and DCAT.
Table~\ref{tab:semantic-error-analysis} summarizes representative semantic failure patterns underlying these results. The models mainly struggle with complex path constraints, such as inverse and sequential paths, and with logical constraints expressed through combinations of \texttt{sh:or} and \texttt{sh:not}. Additional errors include semantic substitutions, such as replacing \texttt{sh:hasValue} with \texttt{sh:class}, and hallucinated vocabulary such as \texttt{sh:condition}.

\begin{table*}[t]
\centering
\caption{Representative semantic error patterns observed in generated SHACL shapes.}
\label{tab:semantic-error-analysis}
\scriptsize
\resizebox{\linewidth}{!}{
\begin{tabular}{p{3.2cm} p{5.8cm} p{5.8cm}}
\toprule
\textbf{Error Type} & \textbf{Reference Pattern} & \textbf{Typical Generated Pattern} \\
\midrule
Inverse Path Omission
& \texttt{sh:path [ sh:inversePath rdf:type ]}
& \texttt{:DatasetShape a sh:NodeShape ;}\newline\texttt{\ \ sh:targetClass dcat:Dataset .}
\\
\midrule
Sequential Path Truncation
& \texttt{sh:path ( rdfs:subClassOf rdf:type )}
& \texttt{sh:path ( rdfs:subClassOf )}
\\
\midrule
Constraint Semantics Substitution
& \texttt{sh:hasValue "http://example.com/BuyerRole" ; sh:path rdf:type}
& \texttt{sh:class ex:BuyerRole}
\\
\midrule
Conditional Constraint Omission
& \texttt{sh:or ( [ sh:not (...) ] [ ... ] )}
& plain property shapes without conditional logic
\\
\midrule
Non-standard Vocabulary Hallucination
& \texttt{sh:or / sh:not}
& \texttt{sh:condition}
\\
\midrule
Datatype Omission
& \texttt{sh:datatype xsd:float}
& datatype constraint missing
\\
\bottomrule
\end{tabular}
}
\end{table*}

\section{Conclusion and Future Work}
\label{sec:conclusion}

In this work, we presented NL2SHACL-Bench, the first benchmark suite specifically designed for natural language to SHACL translation. 
We formally define the NL2SHACL task and introduce a unified framework for dataset construction and evaluation, including semantic metrics based on validation behavior beyond purely structural comparison. 
Our experiments provide a systematic evaluation of current LLMs on this task. 
The results show that while modern LLMs are already highly capable of generating syntactically valid SHACL shapes, substantial challenges remain in producing structurally accurate and semantically equivalent constraints, especially for complex logical patterns and path constraints.
Our experimental results further confirm that the structural metrics and the semantic metrics are complementary to one another. Within the semantic metrics, ASER and LER each have limitations that are disjoint from one another. In practice, this means the choice of metric should be informed by the characteristics of the shape being evaluated, and results across all of them should be cross-analyzed.


Several other directions remain for future work: extending the benchmark with more expressive SHACL features such as SHACL-SPARQL; evaluating NL2SHACL under more practical settings, since our current evaluation assumes exact ontology terms and in-context examples are explicitly provided; and extending the largely constraint-language-independent framework to Shape Expressions (ShEx)~\cite{shex}.

\begin{credits}
\paragraph{Resource Availability Statement:}
The source code for the NL2SHACL-Bench, including the software component and implemented metrics are available on GitHub\footnote{\label{source-code}NL2SHACL-Bench Code: \url{https://github.com/DE-TUM/NL2SHACL-Framework}} and is licensed under the MIT license.
The constructed datasets are publicly available on Zenodo (DOI: 10.5281/zenodo.20082565) under the CC-BY-SA 4.0 license.

\paragraph{Maintainability Statement:}
We plan to actively maintain NL2SHACL-Bench alongside its ongoing development. 
Future updates will include additional datasets, more expressive SHACL features, and improvements to the semantic evaluation pipeline. 
We also plan to publish NL2SHACL-Framework as a PyPI package to lower the barrier for adoption. 
We further aim to extend the framework toward more realistic NL2SHACL settings and additional RDF validation formalisms such as ShEx, ensuring its extensibility and long-term usability for both research and practical applications.

\subsubsection*{Acknowledgements}
This work was supported by the German Research Foundation (DFG) – SFB 1608 – Project-ID 501798263, and DFG - SFB 1625 - 506711657, subproject A06.
Maribel Acosta is supported by a Google Gemini Academic Program Award. 

\end{credits}

\newpage
\section*{Declaration of Use of Generative AI}

Generative AI tools were used to assist with language refinement and parts of the dataset construction process. 
In particular, some initial NL--SHACL pairs were generated with AI assistance and subsequently manually verified and corrected by the authors. 
All scientific content, including the research design, methodology, evaluation, analysis, and final dataset curation, was developed and validated by the authors. 
The authors take full responsibility for the content of this paper and the released resources.


%
%
%
\bibliographystyle{splncs04}
\bibliography{references}

@inproceedings{ahmetaj2025shacl,
  title = {{{SHACL Validation Under Graph Updates}}},
  booktitle = {The {{Semantic Web}} -- {{ISWC}} 2025: 24th {{International Semantic Web Conference}}, {{Nara}}, {{Japan}}, {{November}} 2--6, 2025, {{Proceedings}}, {{Part I}}},
  author = {Ahmetaj, Shqiponja and Konstantinidis, George and Ortiz, Magdalena and Pareti, Paolo and Simkus, Mantas},
  year = 2025,
  month = nov,
  pages = {140--157},
  publisher = {Springer-Verlag},
  isbn = {978-3-032-09526-8}
}

@inproceedings{ahmetaj2025common,
  title={Common foundations for shacl, shex, and pg-schema},
  author={Ahmetaj, Shqiponja and Boneva, Iovka and Hidders, Jan and Hose, Katja and Jakubowski, Maxime and Labra Gayo, Jose Emilio and Martens, Wim and Mogavero, Fabio and Murlak, Filip and Okulmus, Cem and others},
  booktitle={Proceedings of the ACM on Web Conference 2025},
  pages={8--21},
  year={2025}
}

@inproceedings{duan2023xsd2shacl,
  title={XSD2SHACL: capturing RDF constraints from XML schema},
  author={Duan, Xuemin and Chaves-Fraga, David and Dimou, Anastasia},
  booktitle={Proceedings Of The 12th Knowledge Capture Conference 2023},
  pages={214--222},
  year={2023}
}

@inproceedings{duan2024scoop,
  title={SCOOP all the Constraints’ Flavours for your Knowledge Graph},
  author={Duan, Xuemin and Chaves-Fraga, David and Derom, Olivier and Dimou, Anastasia},
  booktitle={European Semantic Web Conference},
  pages={217--234},
  year={2024},
  organization={Springer}
}

@article{pareti2024shacl2fol,
  title={SHACL2FOL: An FOL Toolkit for SHACL Decision Problems},
  author={Pareti, Paolo},
  journal={arXiv preprint arXiv:2406.08018},
  year={2024}
}

@misc{openrouter,
    title = {OpenRouter},
    howpublished = {\url{https://openrouter.ai/rankings?benchmark=coding}},
    note         = {Accessed: 2026/05/07}
}

@misc{shex,
  author       = {Eric Prud'hommeaux and Iovka Boneva and Jose Emilio Labra Gayo and Gregg Kellogg},
  title        = {Shape Expressions Language 2.1},
  howpublished = {W3C Community Group Report},
  month        = oct,
  year         = {2019},
  url          = {https://shex.io/shex-semantics/},
  note         = {Available at \url{https://www.w3.org/ns/shex}}
}

@misc{shacl,
  author       = {{W3C}},
  title        = {Shapes Constraint Language (SHACL)},
  howpublished = {\url{https://www.w3.org/TR/shacl/}},
  note         = {Accessed: 2026/05/07}
}

@misc{chemkgchemrof,
  author       = {{Unknown}},
  title        = {chemkg/chemrof},
  howpublished = {\url{https://github.com/chemkg/chemrof}},
  note         = {Accessed: 2026/05/07}
}

@misc{eprocurementontology,
  author       = {{Unknown}},
  title        = {eProcurement Ontology},
  howpublished = {\url{https://github.com/OP-TED/ePO/tree/master}},
  note         = {Accessed: 2026/05/07}
}

@misc{snikproject,
  author       = {{Unknown}},
  title        = {snikproject},
  howpublished = {\url{https://github.com/snikproject}},
  note         = {Accessed: 2026/05/07}
}

@misc{dcatapshacl,
  author       = {{Unknown}},
  title        = {dcat-ap\_shacl},
  howpublished = {\url{https://github.com/SEMICeu/dcat-ap\_shacl}},
  note         = {Accessed: 2026/05/07}
}

@misc{edifactval,
  author       = {{M{\"a}kelburg, Johannes and John, Christian and Acosta, Maribel}},
  title        = {EDIFACT-VAL},
  howpublished = {\url{https://github.com/DE-TUM/EDIFACT-VAL}},
  note         = {Accessed: 2026/05/07}
}

@inproceedings{jovanovik2025rdfgraphgen,
  title={RDFGraphGen: An RDF Graph Generator Based on SHACL Shapes},
  author={Jovanovik, Milos and Vecovska, Marija and Jakubowski, Maxime and Hose, Katja},
  booktitle={International Joint Conference on Knowledge Graphs},
  pages={111--125},
  year={2025},
  organization={Springer}
}

@inproceedings{auer2007dbpedia,
  title={Dbpedia: A nucleus for a web of open data},
  author={Auer, S{\"o}ren and Bizer, Christian and Kobilarov, Georgi and Lehmann, Jens and Cyganiak, Richard and Ives, Zachary},
  booktitle={international semantic web conference},
  pages={722--735},
  year={2007},
  organization={Springer}
}

@incollection{cimmino2020astrea,
  title = {Astrea: {{Automatic Generation}} of {{SHACL Shapes}} from {{Ontologies}}},
  shorttitle = {Astrea},
  booktitle = {The {{Semantic Web}}},
  author = {Cimmino, Andrea and {Fern{\'a}ndez-Izquierdo}, Alba and {Garc{\'i}a-Castro}, Ra{\'u}l},
  editor = {Harth, Andreas and Kirrane, Sabrina and Ngonga Ngomo, Axel-Cyrille and Paulheim, Heiko and Rula, Anisa and Gentile, Anna Lisa and Haase, Peter and Cochez, Michael},
  year = 2020,
  volume = {12123},
  pages = {497--513},
  publisher = {Springer International Publishing},
  isbn = {978-3-030-49460-5 978-3-030-49461-2},
  langid = {english}
}

@article{evtikhiev2023out,
  title = {Out of the {{BLEU}}: {{How}} Should We Assess Quality of the {{Code Generation}} Models?},
  shorttitle = {Out of the {{BLEU}}},
  author = {Evtikhiev, Mikhail and Bogomolov, Egor and Sokolov, Yaroslav and Bryksin, Timofey},
  year = 2023,
  month = sep,
  journal = {Journal of Systems and Software},
  volume = {203},
  pages = {111741},
  issn = {0164-1212}
}

@article{fernandez-alvarez2022automatic,
  title = {Automatic Extraction of Shapes Using {{sheXer}}},
  author = {{Fernandez-{\'A}lvarez}, Daniel and {Labra-Gayo}, Jose Emilio and {Gayo-Avello}, Daniel},
  year = 2022,
  month = feb,
  journal = {Knowledge-Based Systems},
  volume = {238},
  pages = {107975},
  issn = {0950-7051}
}

@misc{geng2025jsonschemabench,
  title = {{{JSONSchemaBench}}: {{A Rigorous Benchmark}} of {{Structured Outputs}} for {{Language Models}}},
  shorttitle = {{{JSONSchemaBench}}},
  author = {Geng, Saibo and Cooper, Hudson and Moskal, Micha{\l} and Jenkins, Samuel and Berman, Julian and Ranchin, Nathan and West, Robert and Horvitz, Eric and Nori, Harsha},
  year = 2025,
  month = feb,
  number = {arXiv:2501.10868},
  eprint = {2501.10868},
  primaryclass = {cs},
  publisher = {arXiv},
  archiveprefix = {arXiv}
}

@article{liu2024are,
  title = {Are {{LLMs}} Good at Structured Outputs? {{A}} Benchmark for Evaluating Structured Output Capabilities in {{LLMs}}},
  shorttitle = {Are {{LLMs}} Good at Structured Outputs?},
  author = {Liu, Yu and Li, Duantengchuan and Wang, Kaili and Xiong, Zhuoran and Shi, Fobo and Wang, Jian and Li, Bing and Hang, Bo},
  year = 2024,
  month = sep,
  journal = {Information Processing \& Management},
  volume = {61},
  number = {5},
  pages = {103809},
  issn = {0306-4573}
}

@inproceedings{luthfi2022sock,
  title = {{{SoCK}}: {{SHACL}} on {{Completeness Knowledge}}},
  booktitle = {International {{Semantic Web Conference}}},
  author = {Luthfi, Muhammad Jilham and Darari, Fariz and Ashardian, Amanda Carrisa},
  year = 2022,
  langid = {english}
}

@incollection{makelburg2024automation,
  title = {Automation of {{Electronic Invoice Validation Using Knowledge Graph Technologies}}},
  booktitle = {The {{Semantic Web}}},
  author = {M{\"a}kelburg, Johannes and John, Christian and Acosta, Maribel},
  editor = {Mero{\~n}o Pe{\~n}uela, Albert and Dimou, Anastasia and Troncy, Rapha{\"e}l and Hartig, Olaf and Acosta, Maribel and Alam, Mehwish and Paulheim, Heiko and Lisena, Pasquale},
  year = 2024,
  volume = {14664},
  pages = {253--269},
  publisher = {Springer Nature Switzerland},
  isbn = {978-3-031-60625-0 978-3-031-60626-7},
  langid = {english}
}

@inproceedings{rabbani2022shacl,
  title = {{{SHACL}} and {{ShEx}} in the {{Wild}}: {{A Community Survey}} on {{Validating Shapes Generation}} and {{Adoption}}},
  shorttitle = {{{SHACL}} and {{ShEx}} in the {{Wild}}},
  booktitle = {Companion {{Proceedings}} of the {{Web Conference}} 2022},
  author = {Rabbani, Kashif and Lissandrini, Matteo and Hose, Katja},
  year = 2022,
  month = apr,
  pages = {260--263},
  publisher = {ACM},
  isbn = {978-1-4503-9130-6},
  langid = {english}
}

@article{rabbani2023extraction,
  title = {Extraction of {{Validating Shapes}} from {{Very Large Knowledge Graphs}}},
  author = {Rabbani, Kashif and Lissandrini, Matteo and Hose, Katja},
  year = 2023,
  month = jan,
  journal = {Proceedings of the VLDB Endowment},
  volume = {16},
  number = {5},
  pages = {1023--1032},
  issn = {2150-8097},
  langid = {english}
}

@inproceedings{schaffenrath2020benchmark,
  title = {Benchmark for {{Performance Evaluation}} of {{SHACL Implementations}} in {{Graph Databases}}},
  booktitle = {Rules and {{Reasoning}}},
  author = {Schaffenrath, Robert and Proksch, Daniel and Kopp, Markus and Albasini, Iacopo and Panasiuk, Oleksandra and Fensel, Anna},
  editor = {{Guti{\'e}rrez-Basulto}, V{\'i}ctor and Kliegr, Tom{\'a}{\v s} and Soylu, Ahmet and Giese, Martin and Roman, Dumitru},
  year = 2020,
  pages = {82--96},
  publisher = {Springer International Publishing},
  isbn = {978-3-030-57977-7},
  langid = {english}
}

@inproceedings{westermann2025automated,
  title = {Automated {{Validation}} of {{Textual Constraints Against AutomationML}} via {{LLMs}} and {{SHACL}}},
  booktitle = {2025 {{IEEE}} 30th {{International Conference}} on {{Emerging Technologies}} and {{Factory Automation}} ({{ETFA}})},
  author = {Westermann, Tom and K{\"o}cher, Aljosha and Gehlhoff, Felix},
  year = 2025,
  month = sep,
  pages = {1--4},
  issn = {1946-0759}
}

@misc{yang2026structeval,
  title = {{{StructEval}}: {{Benchmarking LLMs}}' {{Capabilities}} to {{Generate Structural Outputs}}},
  shorttitle = {{{StructEval}}},
  author = {Yang, Jialin and Jiang, Dongfu and He, Lipeng and Siu, Sherman and Zhang, Yuxuan and Liao, Disen and Li, Zhuofeng and Zeng, Huaye and Jia, Yiming and Wang, Haozhe and Schneider, Benjamin and Ruan, Chi and Ma, Wentao and Lyu, Zhiheng and Wang, Yifei and Lu, Yi and Do, Quy Duc and Jiang, Ziyan and Nie, Ping and Chen, Wenhu},
  year = 2026,
  month = apr,
  number = {arXiv:2505.20139},
  eprint = {2505.20139},
  primaryclass = {cs},
  publisher = {arXiv},
  archiveprefix = {arXiv}
}

@inproceedings{zhang2025schema,
  title = {Schema {{Generation}} for {{Large Knowledge Graphs Using Large Language Models}}},
  booktitle = {{{EMNLP}}},
  author = {Zhang, Bohui and He, Yuan and Pintscher, Lydia and Pe{\~n}uela, Albert Mero{\~n}o and Simperl, Elena},
  year = 2025,
  langid = {english}
}

\end{document}